\documentclass{article}
\usepackage[preprint]{neurips_2026}

\usepackage[utf8]{inputenc}
\usepackage[T1]{fontenc}
\usepackage{hyperref}
\usepackage{url}
\usepackage{booktabs}
\usepackage{graphicx}
\usepackage{amsmath}
\usepackage{amssymb}
\usepackage{xcolor}
\usepackage{natbib}
\usepackage{float}

\title{Attention Capture Is Not Detection: A Two-Stage Account of How Humans Miss Localized AI Image Edits}

\author{
  Chiao-Chieh Deng \\
  Bachelor's Program in Artificial Intelligence \\
  National Chengchi University, Taipei, Taiwan \\
  \texttt{111702034@g.nccu.edu.tw}
}

\begin{document}

% Small, safe textheight increase, kept well inside the 30pt \footskip
% reserve so the text block cannot overflow past the bottom margin (a much
% larger value was tried earlier and caused real overflow past the page
% edge -- confirmed via the geometry log: adding to \textheight without
% adjusting \topmargin/\footskip pushes the bottom of the text block down
% by the full added amount, not distributed symmetrically). Any remaining
% page-budget gap should be closed with real content/figure trims, not by
% increasing this value further.
\addtolength{\textheight}{12pt}

\maketitle

\begin{abstract}
As AI-generated image edits proliferate, the platforms meant to curb the resulting disinformation treat detectability as a single, undifferentiated property: an edit either gets a warning or it does not. We show this is the wrong model. Across a controlled eye-tracking study (N=59, Latin-square design, four conditions crossing edit area and semantic plausibility), a mixed-effects analysis reveals that whether an edit is \emph{noticed} and whether it is \emph{correctly judged as fake} are dissociable stages, governed by different factors: edit area drives attention capture ($p<0.001$) while semantic plausibility drives judgment accuracy and look-but-fail-to-see (LBFS) error rates ($p<0.001$). This dissociation survives correction for multiple comparisons; a secondary interaction between the two factors does not. This two-stage account extends a long-standing distinction in visual attention research (between pre-attentive capture and effortful recognition) into the new domain of AI-edit detectability. We then test whether a generative eye-movement model can computationally operationalize the attention-capture stage: a Transformer trained to generate scanpaths tracks per-image attention with strong discriminative power (Pearson $r=0.77$--$0.82$ across held-out stimuli) and, on the harder task of predicting LBFS incidence, modestly outperforms a two-parameter linear baseline even without access to the plausibility label ($r=0.52$ vs.\ $0.48$). We report this comparison, our ablations, and our method's limitations (a single fixed train/validation split, not leave-one-subject-out) without inflation, consistent with responsibly communicating what a machine learning system can and cannot do to help curb AI-driven disinformation.
\end{abstract}


\section{Introduction}

Content platforms increasingly attach ``Made with AI'' labels to images edited by generative tools. The policy is coarse: a photographer removing a stray object receives the same label as a fabricated image. This produces two failure modes: trivial edits get flagged as false accusations, eroding trust; and audiences habituate to undifferentiated labels, a well-documented warning-fatigue pattern \citep{bravolillo2014}. Neither is fixed by improving \emph{detection accuracy}; both require answering: \textbf{will a human notice this edit, and if so, correctly recognize it as manipulated?}

Existing manipulation-detection research answers a narrower question: can a classifier determine an image was altered \citep[survey the scope of this literature]{kadha2025}. That is orthogonal to ours: a classifier can be certain an edit exists while telling us nothing about whether a human would ever notice it.

We argue human detectability is not one quantity but two dissociable cognitive stages: an edit must first capture attention, and, independently, an attended edit must be recognized as anomalous. This is not new to cognitive psychology: Mack and Rock's (\citeyear{mackrock1998}) work on inattentional blindness established that perception requires attention; Wolfe's Guided Search formalizes a pre-attentive stage determining where the eyes go, followed by a capacity-limited stage determining what gets recognized \citep{wolfe2021}. These stages dissociate dramatically: 83\% of radiologists searching for lung nodules fixated directly on an inserted gorilla image without reporting it \citep{drew2013}, a canonical look-but-fail-to-see (LBFS) error \citep{wolfe2022}.

Our contribution brings this account to AI-edit detectability:

\begin{enumerate}
\setlength{\itemsep}{1pt}
\setlength{\parskip}{0pt}
\setlength{\topsep}{2pt}
\item \textbf{A statistical demonstration of the two-stage account} (Section~4): edit area and plausibility act on different, dissociable stages (our central, model-independent finding, robust to multiple-comparison correction); a secondary interaction effect is observed but does not survive correction.
\item \textbf{A computational operationalization} (Section~5): a generative scanpath model predicting whether an edit will capture attention ($r=0.77$--$0.82$ against held-out stimuli), with an ablation identifying which components contribute.
\item \textbf{A necessity test against a trivial baseline} (Section~6): a two-parameter linear regression on design-time labels, under leave-one-stimulus-out CV, reporting where the added complexity does and does not earn its keep.
\end{enumerate}

We close with two applications (Section~7) and a limitations section (Section~8) covering webcam eye-tracking precision, sample size, and validation scope.

\section{Related Work}

\textbf{Inattentional blindness and look-but-fail-to-see.} \citet{mackrock1998} established that visual consciousness requires attention, using a paradigm where an unexpected shape goes unreported during an unrelated task. \citet{simons1999} showed roughly half of observers miss a gorilla suit during a counting task. \citet{rensink1997} established the complementary change-blindness paradigm (alternating two photographs differing in one region), the closest methodological precedent to our stimuli, differing mainly in that our changes are AI-generated rather than manually composited. \citet{drew2013} showed domain expertise does not protect against this failure: expert radiologists fixated on an anomalous object without reporting it. \citet{wolfe2022} synthesize this as ``looked-but-failed-to-see'' errors, and Guided Search 6.0 \citep{wolfe2021} provides the formal two-stage account (parallel feature-guided attention followed by serial recognition), adapted as our theoretical scaffold for Section~4.

\textbf{Warning fatigue and AI content labeling.} \citet{bravolillo2014} document that repeated, undifferentiated warnings produce rapid habituation; we treat AI content labels as a structurally similar class, motivating our position that detectability should inform label design.

\textbf{Scanpath generation.} OAT \citep{fang2024} predicts object-level scanpaths over discrete object arrays via encoder--decoder cross-attention. We adopt its cross-attention backbone and follow its evaluation protocol (SS, FED). Our distance-to-AOI encoding (DPE) is inspired by OAT's distance-based positional encoding but is not a direct implementation: it computes distance to a single fixed reference point (the edited region's centre) rather than a learned pairwise Gaussian-decay kernel between arbitrary locations. Our gaze branch's cell positional encoding (CPE) is a standard learned grid-cell embedding, claimed as neither novel nor borrowed. We depart from OAT in domain (continuous, natural-image edits rather than discrete arrays) and, with DiffEye \citep{kara2025} and concurrent ScanDiff \citep{cartella2025}, in modeling gaze probabilistically; both already generate diverse scanpaths via diffusion models. We did not originate the natural-scene setting, probabilistic generation, or distance-encoding, and say so rather than claim contested novelty.

\textbf{Methodology.} Our mixed-effects models use a random-intercept-only structure; \citet{barr2013} is the standard reference for this tradeoff, discussed against their recommendation in Section~4.

\section{Dataset and Experimental Design}

We recruited 59 participants for an online, Latin-square-balanced viewing study. Each viewed 60 base images in one of five conditions: unmodified (O), or AI-edited crossing \textbf{area} (small/large) and \textbf{plausibility} (plausible/implausible): PS, PL, IS, IL. The AOI was defined from the actual inpainting mask rather than a bounding box, which frequently failed to overlap the edited pixels. Participants judged whether each image had been AI-modified, while gaze was recorded continuously.

\textbf{Stimulus generation.} The four variants per base image came from a five-stage automated pipeline (manual authoring at controlled scales for 60 images $\times$ 4 conditions is impractical): (1) GPT-4o describes the scene; (2) an LLM planner proposes a replacement per cell, targeting small (5--10\%) or large (25--40\%) footprints and plausible/implausible fit; (3) GroundingDINO localizes the region (COCO box on failure); (4) a pixel-level mask is generated from the COCO annotation, retried if area falls outside range (this mask defines the AOI); (5) FLUX.1 Fill inpaints per the plan, then compositing restores pixels outside the mask.

A sixth, quality-control stage (5b) guards against a confound: an ``implausible'' edit noticed only for being visibly \emph{low-quality} would be detected for reasons unrelated to implausibility. We run a two-phase blind LLM check per image: phase one asks GPT-4o to identify the modified region \emph{without} the intended edit; phase two checks the match and scores artifacts/coherence. Images failing any of five criteria are regenerated, so detectability differences in Section~4 reflect semantic fit, not rendering quality.

\begin{figure}[H]
\centering
\includegraphics[width=0.64\textwidth]{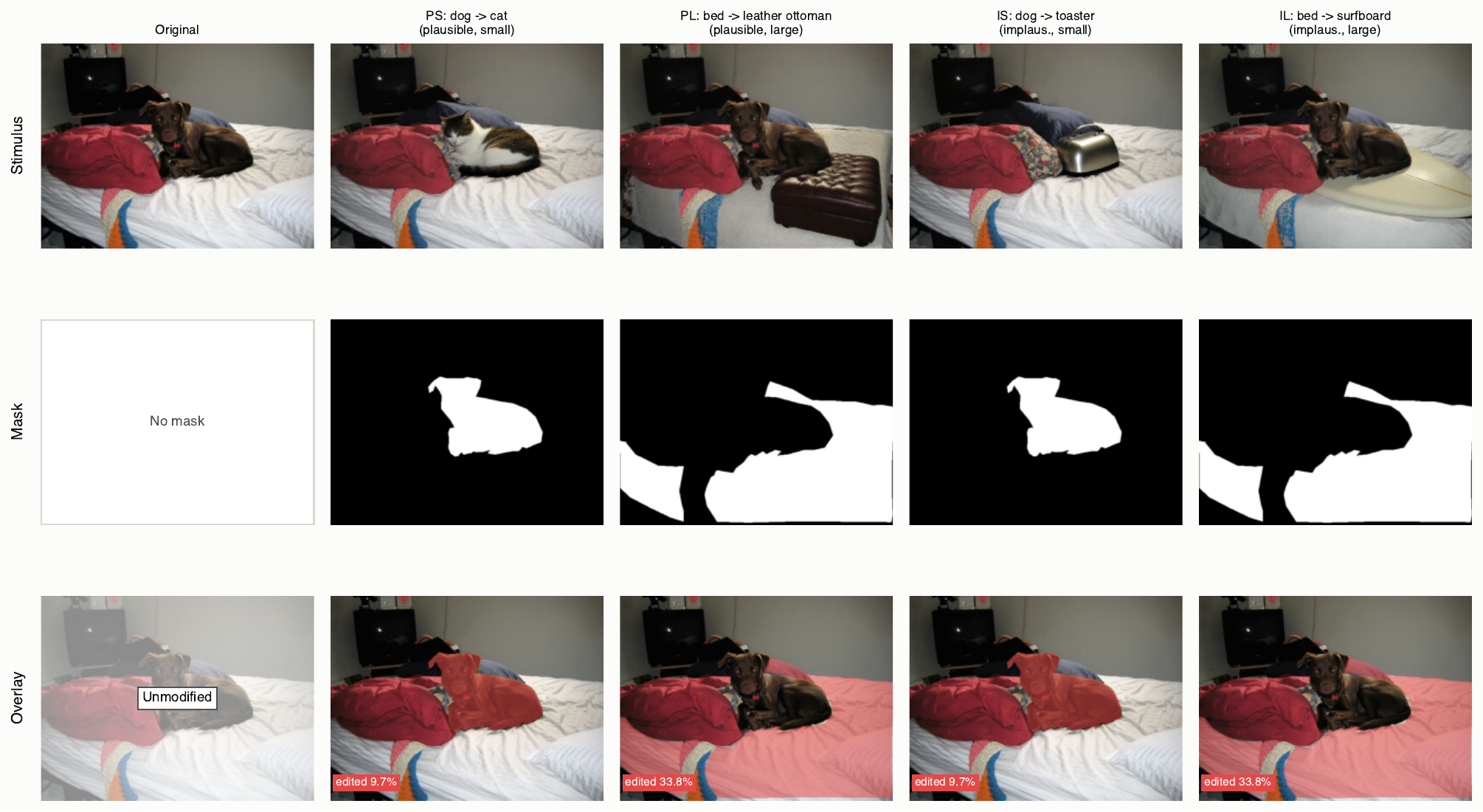}
\caption{One base image rendered across all five conditions (top row), with the corresponding inpainting mask (middle row) and that mask overlaid in translucent red on the unedited original (bottom row, with the edited-area percentage of image pixels). The mask (not an object-detection box) is what defines the AOI used throughout the paper. Object-replacement labels are taken directly from the stimulus-generation pipeline's plan for this image: PS replaces the dog with a cat (9.7\% of image area), PL replaces the bed with a leather ottoman (33.8\%), IS replaces the dog with an implausible toaster (9.7\%), and IL replaces the bed with an implausible surfboard (33.8\%); the plausible/implausible pairs are matched on edit area within each size tier.}
\label{fig:stimdemo}
\end{figure}

Gaze was collected with WebGazer.js \citep{papoutsaki2016}, a browser-based webcam estimator, rather than a lab infrared tracker, so data collection could scale to unsupervised remote participants. This is a precision, not validity, limitation: our statistics rely on coarse dwell/AOI containment; full detail is in Section~8, item 1.

\section{The Two-Stage Detection Model}

This section presents the paper's central, model-independent finding: whether an AI edit is noticed and whether it is correctly identified as fake are governed by different experimental factors, consistent with a two-stage cognitive account rather than a single detectability construct.

\begin{figure}[H]
\centering
\includegraphics[width=0.8\textwidth]{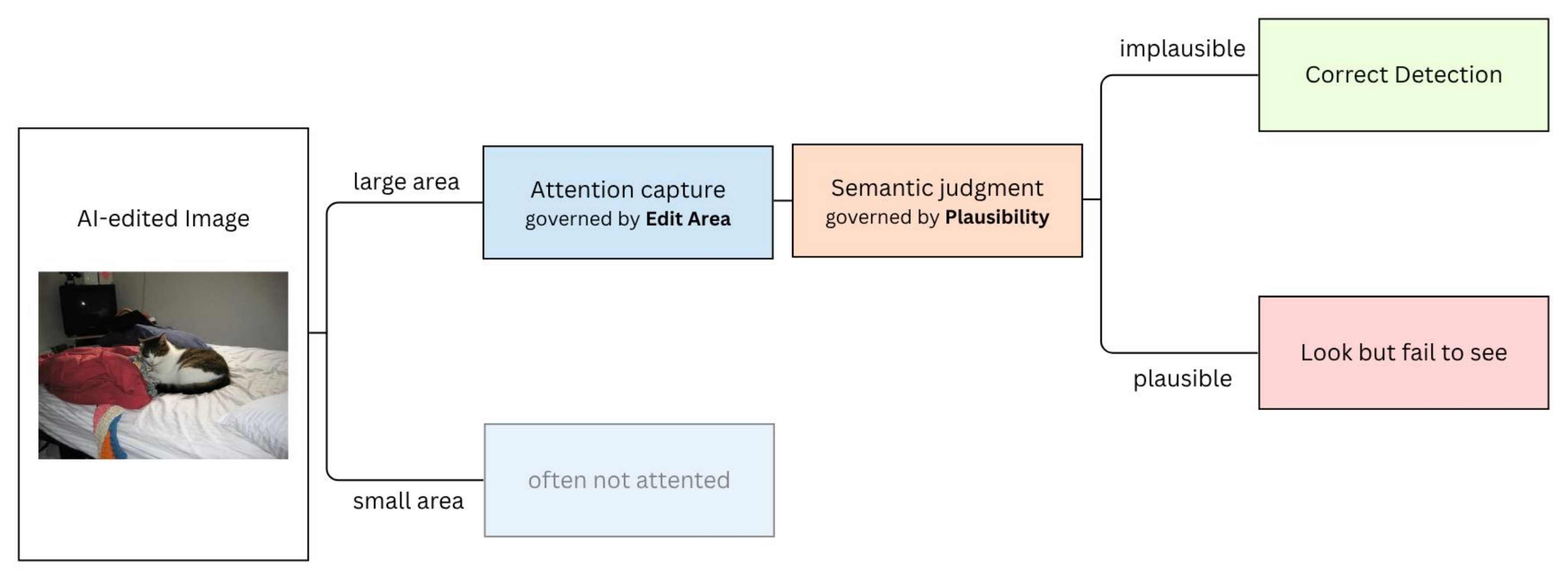}
\caption{An AI edit must first capture attention (governed by edit area) before an attended edit can be correctly judged as fake versus missed as a look-but-fail-to-see error (governed by semantic plausibility). Unattended and unjudged branches are shown in gray.}
\label{fig:schematic}
\end{figure}

We fit generalized linear mixed-effects models of the form \texttt{DV \textasciitilde{} plausibility * size + (1 | participant\_id)} on the modified-condition trials (N=1704 for attention bias and accuracy; N=1381 for LBFS, restricted to trials where the AOI was fixated; N=841 for log-RT, correct trials only; 3109 is the full dataset including unmodified filler trials, which this analysis does not use), for four dependent variables: an attention-bias index, judgment accuracy, LBFS incidence, and log-transformed response time. Following \citet{barr2013}, a maximal structure (crossed by-participant and by-item random intercepts, plus by-item random slopes) would be more conservative for confirmatory testing; our model includes only a by-participant random intercept, omitting both the by-item intercept and any random slopes, for tractability given our sample size. This is anti-conservative and treated as suggestive rather than maximally conservative; it most weakens the interaction terms flagged below as not surviving correction, less so the $p<0.001$ main effects.

The attention-bias index requires definition, because its behavior is easy to misread. It is not raw gaze-dwell proportion; it is dwell proportion \textbf{divided by} the edited region's area as a fraction of the image (\texttt{attention\_bias = dwell\_ratio / area\_fraction}), i.e., dwell relative to the chance level a purely random gaze pattern would produce given the region's size. A value above 1 indicates above-chance attraction; below 1 indicates the opposite.

\begin{table}[H]
\caption{Raw GLMM coefficients, \texttt{DV \textasciitilde{} plausibility * size + (1|participant\_id)}. Accuracy/LBFS use logistic (binomial) family; attention bias/log(RT) use linear (see text). N per row: 1704 (attention bias, accuracy), 1381 (LBFS), 841 (log RT). * $p<.05$, ** $p<.01$, *** $p<.001$ (uncorrected; see Bonferroni discussion below).}
\label{tab:glmm}
\centering
\scriptsize
\begin{tabular}{lccc}
\toprule
Dependent variable & Plausibility effect & Size effect & Interaction \\
 & (implausible vs.\ plausible) & (large vs.\ small) & \\
\midrule
Attention bias (chance-corrected) & $\beta{=}+0.160$, $p{=}0.411$ (n.s.) & $\beta{=}-1.956$, $p{<}0.001$*** & $\beta{=}-0.160$, $p{=}0.559$ (n.s.) \\
Judgment accuracy & $\beta{=}+0.774$, $p{<}0.001$*** & $\beta{=}+0.259$, $p{=}0.011$* & $\beta{=}+0.451$, $p{=}0.027$* \\
LBFS incidence & $\beta{=}-0.611$, $p{<}0.001$*** & $\beta{=}+0.229$, $p{=}0.026$* & $\beta{=}-0.549$, $p{=}0.008$** \\
log(RT) & $\beta{=}-0.066$, $p{=}0.238$ (n.s.) & $\beta{=}-0.012$, $p{=}0.838$ (n.s.) & $\beta{=}-0.057$, $p{=}0.457$ (n.s.) \\
\bottomrule
\end{tabular}
\end{table}

\textbf{Reading the area effect requires two layers, not one.} Raw dwell rises sharply with area (0.155 small vs.\ 0.595 large): large edits capture more \emph{absolute} gaze. But the chance-corrected coefficient above is significantly \emph{negative}: once we divide out that a larger region is mechanically harder to miss by chance, small edits are disproportionately over-attended relative to footprint. Both are true simultaneously: area determines \emph{absolute opportunity} to be seen, not the \emph{density} of attention once size is accounted for.

Critically, area's effect on accuracy and LBFS, while now detectable (below), is far smaller than plausibility's: significant at the raw threshold ($p=0.011$, $p=0.026$) but not after the Bonferroni correction below, unlike plausibility's effect on both ($p<0.001$ for each). Plausibility shows the mirror pattern: no effect on attention capture ($p=0.411$), but the strongest effects in the table on accuracy and LBFS, with a plausibility-by-size interaction on both ($p=0.027$, $p=0.008$). We report raw p-values across twelve tests in Table~1; only three survive Bonferroni correction ($\alpha=0.05/12\approx0.004$): size on attention bias, and plausibility on accuracy and LBFS, exactly the three effects the two-stage account predicts. Area's now-significant (raw-$p$) effects on accuracy/LBFS, and all four interaction terms, do not survive correction. In short: \textbf{plausibility dominates the judgment stage, area dominates attention capture, and this is the reading that survives correction}, a graded, not absolute, version of the dissociation Guided Search predicts \citep{wolfe2021}.

\textbf{Model choice matters here.} These accuracy/LBFS coefficients are logistic; a linear probability model on the same data (an earlier version of this analysis) reported area's effect on both as non-significant ($p=0.816$, $p=0.839$); the choice of family, not the missing by-item random intercept Section~8 discusses, drove that difference. A crossed by-item random intercept shifts the logistic estimates above only marginally (accuracy $p=0.010$; LBFS $p=0.028$).

This anticipates Sections~5--6: generated-gaze exploration reflects only the attention stage, so its correlation with real LBFS should be positive but bounded, since LBFS is co-determined by judgment. That is exactly the pattern we report.

\section{Computational Validation: A Generative Gaze Model}

This section builds the technical component the two label-design applications in Section~7 (graded label exemption, blind-spot highlighting) run on: a model that predicts, per image, whether an edit will draw attention at all.

\subsection{Architecture}

\begin{figure}[H]
\centering
\includegraphics[width=0.92\textwidth]{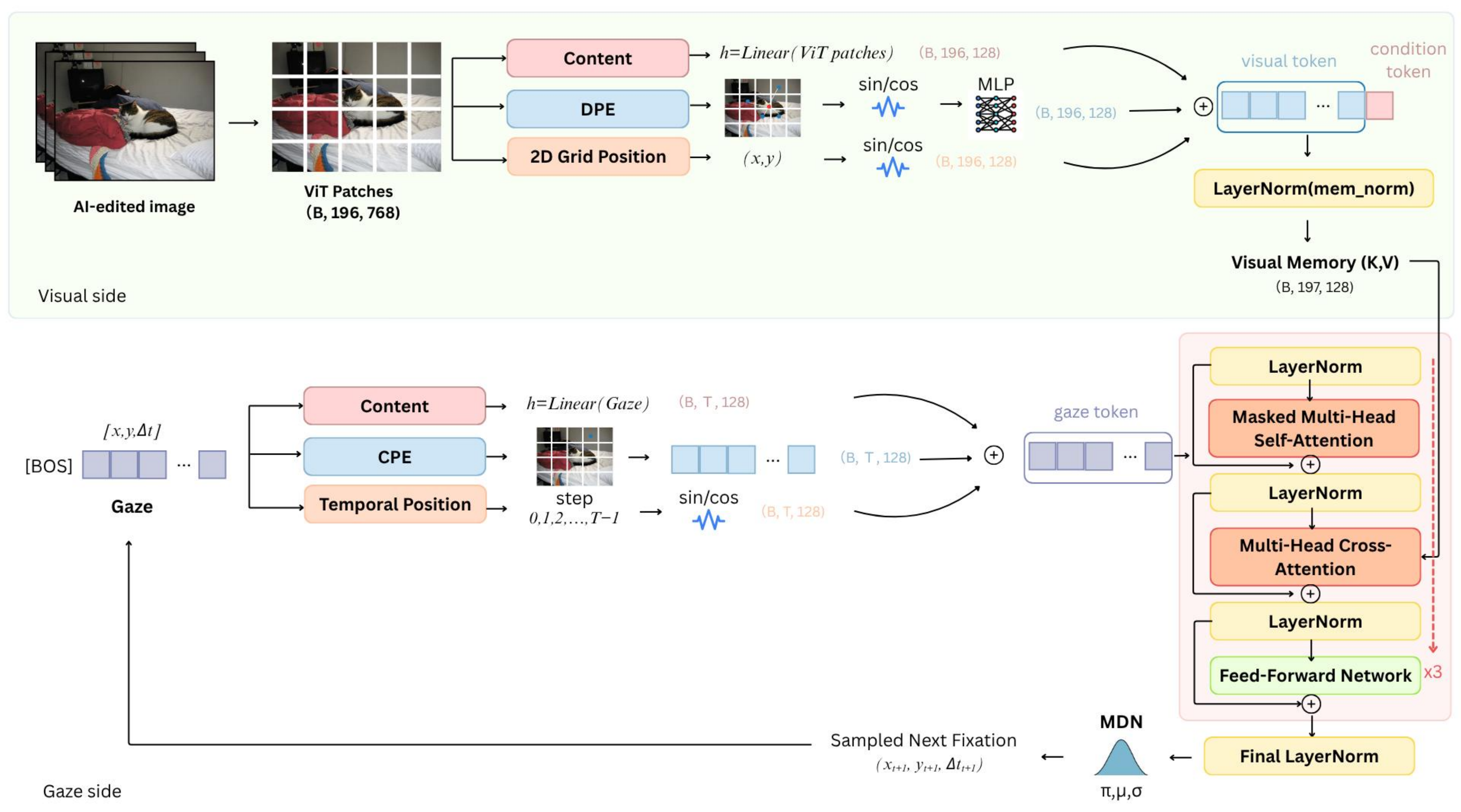}
\caption{SpatioTemporalGazeGen architecture, with tensor shapes at each stage. The visual side (top) combines three parallel branches over ViT patches (B,196,768): content (linear projection), DPE (distance-to-AOI, sinusoidal + MLP), and absolute 2D grid position (sinusoidal); these sum into a visual memory (B,197,128), after concatenating a condition token and layer-normalizing. The gaze side (bottom) combines content, CPE (a direct grid-cell embedding lookup, no sinusoidal step), and a temporal-position sinusoid, then passes through $N{=}3$ Pre-LN Transformer blocks (masked self-attention, cross-attention to the visual memory, feed-forward, each with a residual add) before a final layer norm and a mixture-density head over the next fixation, sampled and fed back in.}
\label{fig:arch}
\end{figure}

We train \textbf{SpatioTemporalGazeGen} (Figure~3), a Transformer \citep{vaswani2017} that generates a full scanpath given an image and its edited region, reading behavioral statistics directly off the trajectory rather than from a separate classification head. DPE, CPE, the block structure, and the mixture-density output head are defined in Figure~3 and Section~2.

One detail matters for validity: recorded gaze is not confined to the frame (12.4\% of samples fall outside $[0,1]$ from webcam noise), so the output maps to $[-0.5, 1.5]$ rather than clipping, avoiding a structural inability to fit real targets.

\subsection{Evaluation protocol and reproducibility}

Results below follow fixes to several evaluation issues found during development (sampling seeds, participant-split determinism, baseline consistency, training-data coverage), averaged over three seeds with a fixed split (Sections~6, 8 discuss not varying it).

\subsection{Main results}

Sequence Score (SS) and Fixation Edit Distance (FED), symbolizing each trajectory as a string over \{inside AOI, near AOI, elsewhere\} against held-out human trajectories, are statistically indistinguishable from the human self-consistency ceiling: SS $= 0.362 \pm 0.016$ vs.\ $0.347$; FED $= 0.638 \pm 0.016$ vs.\ $0.653$: parity, not superiority; the gap is smaller than run-to-run variance. This parity is only informative relative to a floor: a single-seed check against two trivial baselines aligned with OAT's protocol (Random: uniform fixations; Center: Gaussian around image centre) confirms the model is not merely matching the ceiling via bland, population-average paths: Random/Center (0.272/0.728, 0.275/0.725) sit clearly below the model (0.346/0.654), which sits almost exactly at the ceiling.

\begin{figure}[H]
\centering
\includegraphics[width=0.78\textwidth]{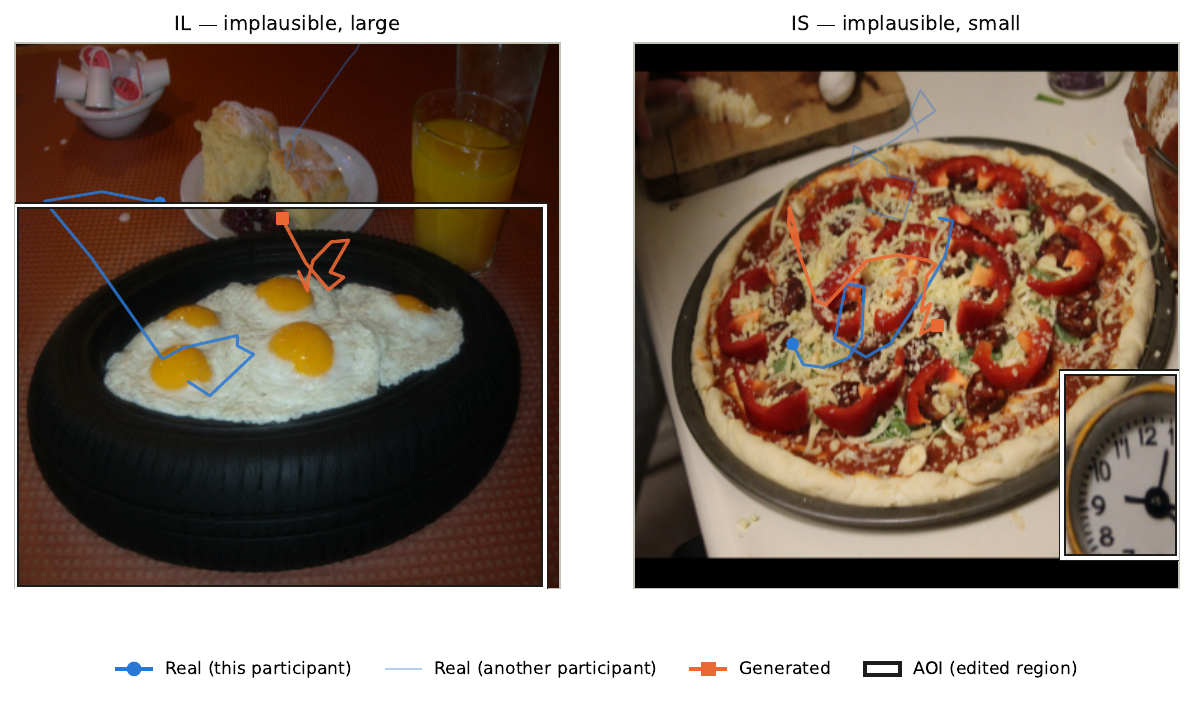}
\caption{Generated vs.\ real scanpaths on two held-out stimuli (temperature 0.7, first 120 fixations; two real participants per panel, to show inter-observer spread). Left, a large edit: real and generated trajectories both enter and linger in the AOI. Right, a small edit (the clock, lower right) that neither the participant nor the model ever visits; both circle the unedited pizza instead. The right panel is the more informative one: reproducing \emph{where humans fail to look} is what Section~7 requires, and is not what a model that merely tracks salient objects would produce.}
\label{fig:scanpaths}
\end{figure}

More informative is discriminative, not aggregate, evidence: per-stimulus Pearson correlation between generated and real exploration is $r=0.77$--$0.82$ across the two best ablation variants: the model correctly ranks which images draw more attention, the property Section~7's application requires. This is not uniform across conditions. In one extraction pass (pooled $r=0.74$, $n=163$), the within-condition correlations are PS $r=0.63$ [95\% CI $0.41$, $0.78$], PL $0.52$ [$0.26$, $0.71$], IS $0.44$ [$0.14$, $0.67$], IL $0.44$ [$0.13$, $0.66$] (Fisher $z$; $n=36$--$45$ per cell, hence the wide intervals). Part of the pooled figure therefore reflects correctly separating small- from large-edit clusters rather than fine-grained within-condition ranking. Both are real, but pooled $r$ overstates within-condition power, so we report the split.

Population-variability calibration (spread of K sampled trajectories vs.\ real participants) is close to 1.0 for exploration, first-entry, and dwell run (ratios $0.81$--$1.09$), indicating the mixture-density design represents disagreement rather than collapsing to one path. Re-entries are consistently weak across every measure: largest rel.diff, lowest correlation ($r\approx0.24$--$0.49$), poorest calibration (ratio $0.70$); discussed mechanistically below.

\subsection{Ablation: which components actually matter}

We ablate five architectural choices against the full model, holding data, seed, and protocol fixed. This single-seed table's ``Full'' row SS ($0.371$) differs slightly from the three-seed headline ($0.362\pm0.016$); both are consistent with the ceiling ($0.347$) within sampling variance.

\begin{table}[H]
\caption{Ablation study. $|$rel.diff$|$ columns ($^\delta$) are absolute relative difference from human behavior on that statistic, closer to 0 = better; sign of the source values is dropped for readability. Single fixed seed=42 for controlled comparison.}
\label{tab:ablation}
\centering
\scriptsize
\begin{tabular}{lcccccc}
\toprule
Model & SS$\uparrow$ & FED$\downarrow$ & Exploration$^\delta$ & First-entry$^\delta$ & Re-entry$^\delta$ & Dwell-run$^\delta$ \\
\midrule
Full & 0.371 & 0.629 & +0.118 & +0.050 & $-$0.306 & +0.269 \\
$-$ condition token & 0.372 & 0.628 & +0.115 & +0.040 & $-$0.263 & +0.248 \\
$-$ cross-attention & 0.361 & 0.639 & +0.121 & +0.070 & $-$0.274 & +0.277 \\
$-$ CPE & 0.374 & 0.626 & \textbf{+0.313} & +0.030 & \textbf{$-$0.557} & \textbf{+0.588} \\
$-$ DPE & 0.362 & 0.638 & +0.077 & +0.083 & $-$0.309 & +0.235 \\
Deterministic (no MDN) & \textbf{0.326} & \textbf{0.674} & \textbf{+0.453} & \textbf{+0.161} & $-$0.131 & \textbf{+0.909} \\
\bottomrule
\end{tabular}
\end{table}

Only two ablations degrade behavior substantially: removing CPE more than doubles exploration/dwell-run error and nearly doubles re-entry error, barely moving SS/FED: its contribution is to \emph{how the model dwells}, not sequence ordering. Removing the MDN head degrades almost every metric, confirming distributional modeling is the second load-bearing choice. The rest (condition token, cross-attention, DPE) contribute little; removing the condition token even \emph{improves} re-entry $r$ ($0.24\to0.49$), read as redundancy: area is already implicit in DPE's AOI geometry, and plausibility barely affects exploration.

\section{Does the Complexity Earn Its Keep? A Baseline Comparison}

The applications in Section~7 rest on the model correctly ranking \emph{which} images will be noticed or missed, not just matching population averages. We test it against the simplest baseline: a two-parameter linear regression on edit area and plausibility (design-time labels, no gaze/vision/network) fit with leave-one-stimulus-out CV so the trivial model gets no unfair in-sample advantage.

\begin{table}[H]
\caption{Trivial-baseline comparison for predicting real LBFS incidence rate (leave-one-stimulus-out, n=163).}
\label{tab:baseline}
\centering
\scriptsize
\begin{tabular}{lc}
\toprule
Predictor & Correlation with real LBFS incidence rate \\
\midrule
Linear(area, plausibility), LOOCV & $r = 0.48$ \\
Generated exploration (full model, with plausibility label) & $r = 0.55$ \\
Generated exploration (condition token removed, \textbf{no plausibility label}) & $r = 0.52$ \\
\bottomrule
\end{tabular}
\end{table}

The full model modestly outperforms the linear baseline, and, more informatively, so does the variant given no explicit plausibility label, relying only on raw pixels and edited-region geometry. This suggests the generated scanpath carries detectability information beyond the two design-time labels, a small but real point favoring vision over a lookup table.

Two qualifications apply. First, both models share the same fixed participant split; the baseline is cross-validated within it, but the generative model has not been re-estimated under a different \emph{participant} split, so the margin may be optimistic relative to full LOSO (Section~8). Second, the comparison is narrow: it scores a scalar risk value, not the spatial trajectory blind-spot highlighting (Section~7.2) requires and which no scalar baseline can provide by construction.

\section{Applications}

\textbf{7.1 Graded label exemption.} Predicted exploration (the strongest-validated output, Section~5.3) could filter edits: those drawing negligible attention become candidates for label exemption, reducing creator friction and warning fatigue \citep{bravolillo2014}.

\textbf{7.2 Blind-spot highlighting.} Where a platform warns, the same trajectory could drive an interface highlighting the region most likely overlooked, rather than undifferentiated text (unavailable to any scalar classifier).

Neither application has been deployed or user-tested; both follow directly from Section~5.3, not evaluated interface designs.

\section{Limitations}

\begin{enumerate}
\setlength{\itemsep}{1pt}
\setlength{\parskip}{0pt}
\setlength{\topsep}{2pt}
\item \textbf{Data collection precision.} WebGazer.js \citep{papoutsaki2016} rather than a lab tracker (precision $\sim$1--1.4\textdegree{}; online replications show 20--30\% attenuation \citep{steffan2024}), and per-sample timestamps were unavailable for all 3{,}280 trials, so times are evenly interpolated. Claims operate at AOI-level dwell scale, robust to this; time-based statistics are ordinal, not measured latency.
\item \textbf{Sample size and validation split.} N=59 is pilot-scale; validation has roughly 11--12 participants, and SS/FED ceilings are often from only 2--3 viewers per stimulus. Results use a single fixed split rather than LOSO (Section~6 discusses the consequence).
\item \textbf{Re-entries are a persistent weakness} (worst on every metric tried), contributing little to predicting real LBFS ($r\approx0.03$--$0.09$), consistent with re-entry reflecting a detection component this architecture does not capture.
\end{enumerate}

\section{Conclusion}

Human detection of AI-edited images is not one phenomenon but two: attention capture, governed by edit area, and judgment, governed by semantic plausibility, statistically dissociated in a study surviving multiple-comparison correction. A generative gaze model operationalizes the attention half with strong per-image discriminative power, and offers a modest, honestly-qualified advantage over a trivial baseline on the harder judgment-adjacent task; its main value is producing a spatial signal no scalar baseline can, not besting that baseline on accuracy. The largest open question is closing the leave-one-subject-out validation gap with laboratory-grade eye tracking. For the labeling policies that motivated this work, the practical upshot is that detectability is not a single dial: a platform that wants fewer false accusations and less warning fatigue needs to know separately whether an edit will be seen at all and, if seen, whether it will be correctly read as fake.


\begin{thebibliography}{99}

\bibitem[Barr et al.(2013)]{barr2013}
Barr, D. J., Levy, R., Scheepers, C., \& Tily, H. J. (2013). Random effects structure for confirmatory hypothesis testing: Keep it maximal. \emph{Journal of Memory and Language, 68}(3), 255--278. \url{https://doi.org/10.1016/j.jml.2012.11.001}

\bibitem[Bravo-Lillo et al.(2014)]{bravolillo2014}
Bravo-Lillo, C., Cranor, L., Komanduri, S., Schechter, S., \& Sleeper, M. (2014). Harder to ignore? Revisiting pop-up fatigue and approaches to prevent it. In \emph{Proceedings of the Tenth Symposium on Usable Privacy and Security (SOUPS 2014)} (pp. 105--111). USENIX Association.

\bibitem[Cartella et al.(2025)]{cartella2025}
Cartella, G., Cuculo, V., D'Amelio, A., Cornia, M., Boccignone, G., \& Cucchiara, R. (2025). Modeling human gaze behavior with diffusion models for unified scanpath prediction. In \emph{Proceedings of the IEEE/CVF International Conference on Computer Vision (ICCV 2025)}. \url{https://openaccess.thecvf.com/content/ICCV2025/papers/Cartella_Modeling_Human_Gaze_Behavior_with_Diffusion_Models_for_Unified_Scanpath_ICCV_2025_paper.pdf} [no DOI assigned at time of writing]

\bibitem[Drew et al.(2013)]{drew2013}
Drew, T., V\~{o}, M. L.-H., \& Wolfe, J. M. (2013). The invisible gorilla strikes again: Sustained inattentional blindness in expert observers. \emph{Psychological Science, 24}(9), 1848--1853. \url{https://doi.org/10.1177/0956797613479386}

\bibitem[Fang et al.(2024)]{fang2024}
Fang, Y., Yu, J., Zhang, H., van der Lans, R., \& Shi, B. (2024). OAT: Object-level attention transformer for gaze scanpath prediction. In \emph{Proceedings of the European Conference on Computer Vision (ECCV 2024)}.

\bibitem[Kadha et al.(2025)]{kadha2025}
Kadha, V., Bakshi, S., \& Das, S. (2025). Unravelling digital forgeries: A systematic survey on image manipulation detection and localization. \emph{ACM Computing Surveys}. \url{https://doi.org/10.1145/3731243}

\bibitem[Kara et al.(2025)]{kara2025}
Kara, O., Nisar, H., \& Rehg, J. M. (2025). DiffEye: Diffusion-based continuous eye-tracking data generation conditioned on natural images. \emph{arXiv:2509.16767}.

\bibitem[Mack and Rock(1998)]{mackrock1998}
Mack, A., \& Rock, I. (1998). \emph{Inattentional blindness}. MIT Press.

\bibitem[Papoutsaki et al.(2016)]{papoutsaki2016}
Papoutsaki, A., Sangkloy, P., Laskey, J., Daskalova, N., Huang, J., \& Hays, J. (2016). WebGazer: Scalable webcam eye tracking using user interactions. In \emph{Proceedings of the 25th International Joint Conference on Artificial Intelligence (IJCAI 2016)}.

\bibitem[Rensink et al.(1997)]{rensink1997}
Rensink, R. A., O'Regan, J. K., \& Clark, J. J. (1997). To see or not to see: The need for attention to perceive changes in scenes. \emph{Psychological Science, 8}(5), 368--373. \url{https://doi.org/10.1111/j.1467-9280.1997.tb00427.x}

\bibitem[Simons and Chabris(1999)]{simons1999}
Simons, D. J., \& Chabris, C. F. (1999). Gorillas in our midst: Sustained inattentional blindness for dynamic events. \emph{Perception, 28}(9), 1059--1074. \url{https://doi.org/10.1068/p281059}

\bibitem[Steffan et al.(2024)]{steffan2024}
Steffan, A., Zimmer, L., et al. (2024). Validation of an open source, remote web-based eye-tracking method (WebGazer) for research in early childhood. \emph{Infancy, 29}(1), 31--55. \url{https://doi.org/10.1111/infa.12564} [32-author multi-lab consortium study; full author list verified but abbreviated here per convention]

\bibitem[Vaswani et al.(2017)]{vaswani2017}
Vaswani, A., Shazeer, N., Parmar, N., Uszkoreit, J., Jones, L., Gomez, A. N., Kaiser, {\L}., \& Polosukhin, I. (2017). Attention is all you need. In \emph{Advances in Neural Information Processing Systems (NeurIPS 2017)}.

\bibitem[Wolfe(2021)]{wolfe2021}
Wolfe, J. M. (2021). Guided Search 6.0: An updated model of visual search. \emph{Psychonomic Bulletin \& Review, 28}(4), 1060--1092. \url{https://doi.org/10.3758/s13423-020-01859-9}

\bibitem[Wolfe et al.(2022)]{wolfe2022}
Wolfe, J. M., Kosovicheva, A., \& Wolfe, B. (2022). Normal blindness: When we look but fail to see. \emph{Trends in Cognitive Sciences, 26}(9), 809--819. \url{https://doi.org/10.1016/j.tics.2022.06.006}

\end{thebibliography}
\end{document}